\pdfoutput=1

\documentclass[11pt]{article}

\usepackage{acl}

\usepackage{times}
\usepackage{latexsym}
\usepackage{graphicx}
\usepackage{algorithm,algorithmic}
\usepackage{multirow}
\usepackage{amsmath} 
\usepackage{bm}
\usepackage{pxrubrica}

\newcommand{\CodeFn}[1]{\texttt{Cd}(#1)}
\newcommand{\IdFn}[1]{\texttt{Id}(#1)}

\usepackage[T1]{fontenc}

\usepackage[utf8]{inputenc}

\usepackage{microtype}

\usepackage{subfig}
\usepackage{CJKutf8}

\newcommand\blfootnote[1]{%
  \begingroup
  \renewcommand\thefootnote{}\footnote{#1}%
  \addtocounter{footnote}{-1}%
  \endgroup
}

\newcommand{\Ja}[1]{\begin{CJK}{UTF8}{ipxm}#1\end{CJK}}

\title{Vaporetto: Efficient Japanese Tokenization Based on Improved Pointwise Linear Classification$^*$}

\author{Koichi Akabe$^1$ \hspace{2em} Shunsuke Kanda$^1$ \hspace{2em} Yusuke Oda$^{1,2}$ \hspace{2em} Shinsuke Mori$^3$ \\
  $^1$ LegalForce Research, Japan
  $^2$ Tohoku University, Japan
  $^3$ Kyoto University, Japan \\
  \texttt{\{vbkaisetsu,shnsk.knd\}@gmail.com} \\
  \texttt{yusuke.oda@predicate.jp},
  \texttt{forest@i.kyoto-u.ac.jp}}

\begin{document}
\maketitle

\rubysizeratio{0.65}
\rubyintergap{-0.30}

\begin{abstract}
This paper proposes an approach to improve the runtime efficiency of Japanese tokenization based on the pointwise linear classification (PLC) framework, which formulates the whole tokenization process as a sequence of linear classification problems. Our approach optimizes tokenization by leveraging the characteristics of the PLC framework and the task definition. Our approach involves (1) composing multiple classifications into array-based operations, (2) efficient feature lookup with memory-optimized automata, and (3) three orthogonal pre-processing methods for reducing actual score calculation. Thus, our approach makes the tokenization speed 5.7 times faster than the current approach based on the same model without decreasing tokenization accuracy. Our implementation is available at \url{https://github.com/daac-tools/vaporetto} under the MIT or Apache-2.0 license.\blfootnote{$^*$They conducted this study while affiliated with the organizations listed.}
\end{abstract}

\section{Introduction}
\label{sec:introduction}

\begin{figure}[t]
    \centering
    \includegraphics[width=0.95\hsize]{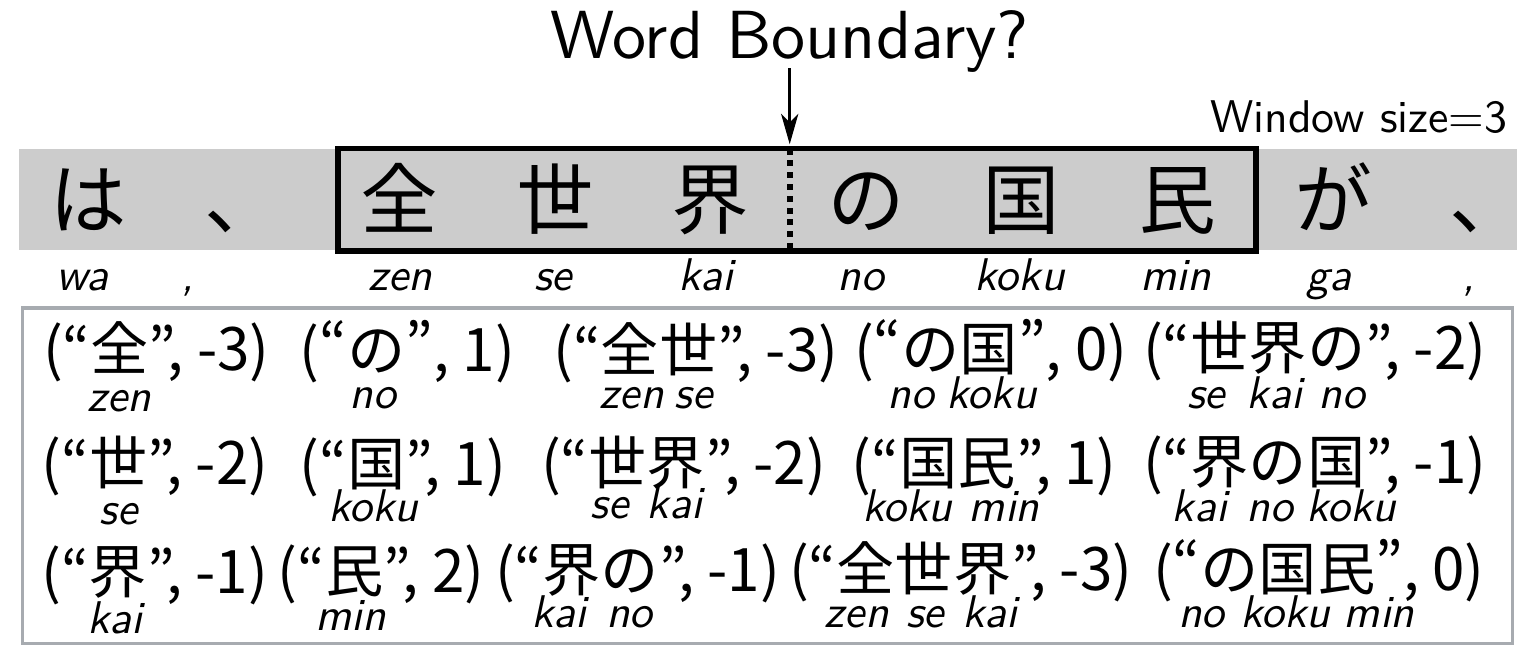}
    \caption{Example of Japanese tokenization with pointwise method.
    Bottom box contains character $n$-gram features described in Section \ref{sec:context-features}.
    }
    \label{fig:features}
\end{figure}

In languages without explicit word boundaries, such as Japanese and Chinese, natural language processing systems must determine these boundaries from unsegmented texts before any word-based analyses.
In Japanese processing, two types of linguistic\footnote{There are also unsupervised tokenization methods such as SentencePiece \cite{kudo-richardson-2018-sentencepiece} and its fast implementation \cite{song-etal-2021-fast}, but we focused on linguistic tokenization, which is still essential for word-sensitive methods such as information retrieval.} tokenization methods have been proposed:
\textit{lattice} and \textit{pointwise}.
Lattice methods \cite{hisamitsu1990morph} generate a lattice of possible tokenizations over the input text and determine the path on the lattice that minimizes a given cost function.
In contrast, pointwise methods \cite{10.5555/1764967.1765101,sassano-2002-empirical,neubig-etal-2011-pointwise,abdelali-etal-2016-farasa,kitagawa-komachi-2018-long,tolmachev-etal-2019-shrinking} use a binary classifier to predict whether a particular character boundary becomes a word boundary, as in Figure \ref{fig:features}.
Pointwise methods formulate the segmentation model for each character boundary independently, allowing for efficient domain adaptation through partial annotation \cite{neubig-etal-2011-pointwise,mori2011pointwise}.

Although pointwise methods can work in a linear time of input lengths, as described in Section \ref{sec:pointwise-linear}, they require careful design of the overall algorithm to reduce unnecessary bottlenecks, which is the primary concern of runtime efficiency.
Comparing well-known examples of both types of methods, we can see that KyTea \cite{neubig-etal-2011-pointwise} (a pointwise method) was almost 2.1 times slower than MeCab \cite{kudo-etal-2004-applying} (a lattice method) in our experiments (Section \ref{sec:ex::comp}), indicating that there is still room for improvement.

In this paper, we focus on designing an efficient algorithm of the pointwise methods with linear classifiers, which we call \textit{Pointwise Linear Classification} (PLC), without changing the model structure.
Our algorithm works the same as KyTea but is much faster than the original implementation.
To this end, we (1) formulate the whole PLC algorithm as a set of array manipulations (Section \ref{sec:pattern-match-in-plc}), (2) introduce an efficient pattern matching algorithm to look up features from the text (Section \ref{sec:algo::matching}), and (3) propose three preprocessing methods to reduce runtime score calculation (Sections \ref{sec:algo::char} to \ref{sec:algo::type}).
Experiments show that combining our approaches improves the tokenization speed and eventually runs 5.7 times faster than KyTea in a controlled environment.
We also provide thorough analyses of the proposed methods to capture the tendency of their behavior from different perspectives.

While we focused on speeding up tokenization without changing the tokenization result, another line of research proposes to speed up the downstream task by changing tokens \cite{hofmann-etal-2022-embarrassingly}.
\section{Pointwise Linear Classification}
\label{sec:pointwise-linear}

\subsection{Algorithm Overview}
\label{sec:pointwise}

Pointwise methods use a binary classifier with \textit{context features} to predict whether a particular character boundary becomes a word boundary \cite{10.5555/1764967.1765101}.
\newcite{sassano-2002-empirical} and \newcite{neubig-etal-2011-pointwise} introduced a support vector machine (SVM) with three types of context features: \textit{character $n$-grams}, \textit{type $n$-grams}, and \textit{dictionary features}.
These features are generated within a sliding window of size $W$, which contains a sequence of surrounding characters around the boundary.
Since the classifier is defined independently from the context features,
we can use off-the-shelf binary classification models.
KyTea introduces a linear SVM as the classification model and uses LIBLINEAR \cite{10.5555/1390681.1442794} for training the model parameters.
Based on these characteristics, KyTea models can be considered PLC variants. As discussed later, PLC models are capable of many optimization techniques.

A single classification in a PLC model is formulated as follows:
\begin{align}
y_i(\bm{w}, x_i) := \bm{w}^\mathsf{T} \bm{\phi}(x_i) + b, \label{eq:linear}
\end{align}
where $x_i$ is the $i$-th input data corresponding to the $i$-th character boundary, $\bm{\phi}(x_i)$ is a binary vector representing a set of available features extracted from $x_i$, $\bm{w}$ is a weight vector corresponding to all features, and $b$ is a scalar bias.
The resulting value $y_i$ suggests the likelihood of the character boundary becoming a word boundary; the higher it is, the more likely it becomes.
When $y_i = 0$ is the decision boundary of this formulation, the classifier determines the word boundary where $y_i$ is positive.

\subsection{Context Features}
\label{sec:context-features}

As shown in Figure \ref{fig:features}, we input a piece of Japanese text, and the model predicts whether the character boundary ``\Ja{\ruby[S]{界}{\textit{kai}}}-\Ja{\ruby[S]{の}{\textit{no}}}'' is a word boundary.
Figure \ref{fig:features} shows available character $n$-gram features in this prediction.
The \textit{character n-gram features} are defined as a pair of its string and the relative position from the corresponding boundary.

\begin{figure}[t]
    \centering
    \includegraphics[width=0.95\hsize]{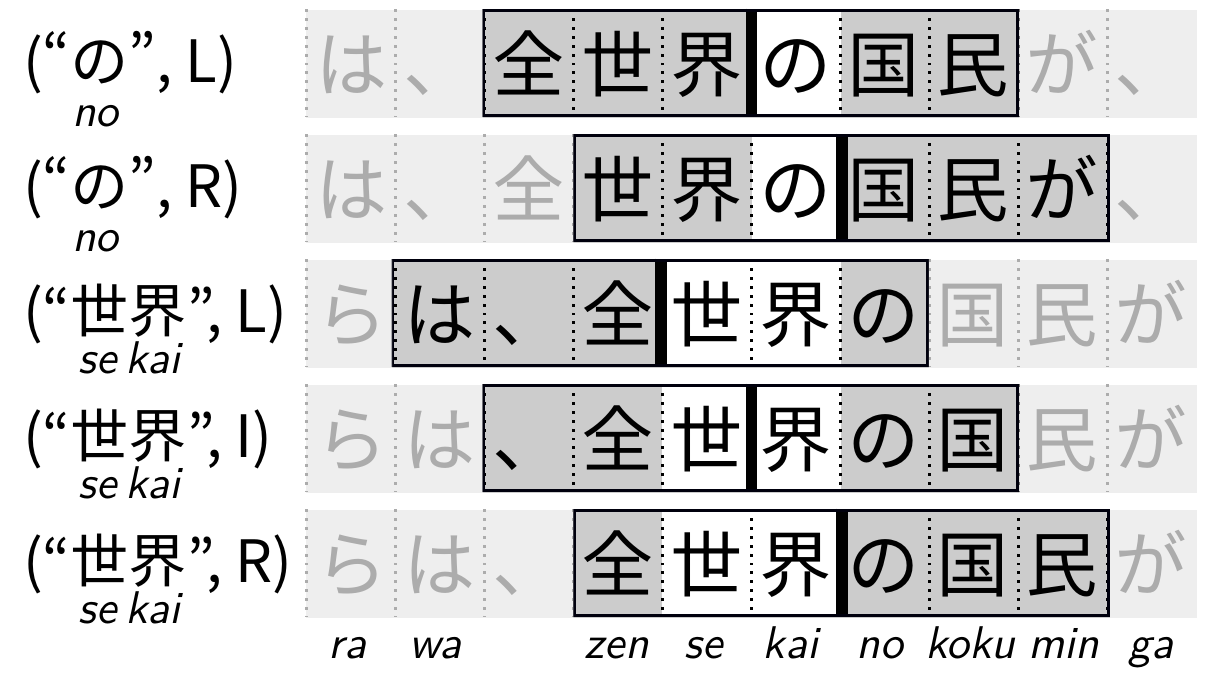}
    \caption{Examples of dictionary features of two words ``\Ja{\ruby[S]{の}{\textit{no}}}''  and ``\Ja{\ruby[S]{世}{\textit{se}}\ruby[S]{界}{\textit{kai}}}'' with different positions.
    Highlighted rectangles indicate position where dictionary word was found. Bold lines indicate position the corresponding feature affects. Solid-line rectangles containing 6 characters indicate window.
    I feature is repeated for all intermediate character boundaries.}
    \label{fig:dict-feature}
\end{figure}

Similar to character $n$-grams, the \textit{type $n$-gram features} are defined as a pair of \textit{character type} $n$-grams and their relative positions. Character types are defined as a function $t(a)$ that assigns a character $a$ to one of six categories: H (Hiragana), T (Katakana), K (Kanji), D (Digit), R (Roman), and O (Other).
For example, the character $n$-gram ``\Ja{\ruby[S]{世}{\textit{se}}\ruby[S]{界}{\textit{kai}}\ruby[S]{の}{\textit{no}}}'' is mapped to the type $n$-gram $[t($`\Ja{\ruby[S]{世}{\textit{se}}}'$), t($`\Ja{\ruby[S]{界}{\textit{kai}}}'$), t($`\Ja{\ruby[S]{の}{\textit{no}}}'$)]=[$`K'$,$`K'$,$`H'$]$.

If some dictionary words overlap a character boundary, the \textit{dictionary features} corresponding to the word are additionally introduced to enhance the confidence of the prediction.\footnote{According to \newcite{neubig-etal-2011-pointwise}, word features are shared among all words with the same number of characters. Our definition is more general as well as covering the original definition.}
Each dictionary word has at most three types of features: L (the leftmost side of the word), R (the rightmost side of the word), and I (any boundaries inside the word).
Figure \ref{fig:dict-feature} shows an example of these dictionary features.

\subsection{PLC with Pattern Matching}
\label{sec:pattern-match-in-plc}

In Equation (\ref{eq:linear}), we must extract the features $\bm{\phi}(x_i)$ and calculate an inner product for every character boundary. Since consecutive classifications in PLC require similar features, this process can be composed into a unified routine in Algorithm \ref{alg:fast-pointwise-prediction}.

Figure \ref{fig:weightvec} shows an example of this algorithm with character $n$-gram features.
First, the input text of $N$ characters is decomposed into a character sequence $\bm{a} = (a_0, a_1, \dots, a_{N-1})$ and analyzed using the pattern-matching function Match$(\bm{a})$ to look up all character $n$-grams with non-zero weights.
The \textit{score array} $\bm{w}_{\textrm{pattern}}(\bm{q})$ is defined for each $n$-gram pattern $\bm{q} \in \text{Match}(\bm{a})$ and formulated as follows:
\begin{align}
&\bm{w}_{\textrm{pattern}}(\bm{q}) \nonumber \\
&:= (w_{(\bm{q},W-n)}, w_{(\bm{q},W-n-1)},..., w_{(\bm{q},-W)}), \label{eq:w-pattern}
\end{align}
where $w_{(\bm{q},\cdot)}$ is a weight corresponding to the $n$-gram feature $(\bm{q},\cdot)$ described in Section \ref{sec:context-features} and $n$ is the length of the $n$-gram. Each score array contains $2W-n+1$ elements corresponding to all relative positions.
For each $\bm{q}$, a corresponding score array is integrated to the appropriate span of the \textit{result array} $(y_1, y_2, ..., y_{N-1})$, representing a sequence of classification results.

\begin{algorithm}[t]
    \caption{PLC using $n$-gram pattern matching}
    \label{alg:fast-pointwise-prediction}
    \begin{algorithmic}[1]
        \renewcommand{\algorithmicrequire}{\textbf{Input:}}
        \renewcommand{\algorithmicensure}{\textbf{Output:}}
        \REQUIRE Text: $\bm{a}$, Weights: $\bm{w}$
        \ENSURE Result array: $(y_1, y_2, \dots,y_{N-1})$
        \STATE $(y_1, y_2, \dots,y_{N-1}) \longleftarrow \bm{0}$
        \FOR {$[j, k) \in $ Match$(\bm{a})$}
            \STATE $\bm{q} \longleftarrow (a_j, a_{j+1}, \dots, a_{k-1})$
            \STATE $\bm{w}' \longleftarrow \bm{w}_{\textrm{pattern}}(\bm{q})$
            \STATE $p \longleftarrow k-W$
            \FOR {$i = 0$ to $2W-n$}
                \IF {$p+i \in [1,N-1]$}
                    \STATE $y_{p+i} \longleftarrow y_{p+i} + w'_i$
                \ENDIF
            \ENDFOR
        \ENDFOR
        \RETURN $(y_1, y_2, \dots,y_{N-1})$ 
    \end{algorithmic} 
\end{algorithm}

Although Algorithm \ref{alg:fast-pointwise-prediction} does not show any improvements in complexity, the calculation of $y_i$ is decomposed into elementwise summing between multiple arrays that can bring high hardware-level throughput.\footnote{Specifically, arithmetic with sequential access is beneficial for (i) accurate branch prediction, (ii) high availability of hardware caches, and (iii) high availability of SIMD optimizations.}

\paragraph{Bottlenecks in KyTea}

KyTea is a straightforward implementation of Algorithm 1, and there is room for several improvements in time efficiency.
One is Match$(\bm{a})$ (line 2).
Another is the process of calculating scores (line 8).
Section \ref{sec:algo} describes the bottlenecks and introduces methods for improving time efficiency.
\section{Improving Efficiency of PLC}
\label{sec:algo}

\subsection{Efficient Pattern Matching}
\label{sec:algo::matching}

The Match$(\bm{a})$ runs over the whole input text $\bm{a}$ and discovers all available substrings registered in the pattern set of the function. We need three pattern-matching functions to achieve the full feature lookup: character $n$-grams, type $n$-grams, and dictionary. The character $n$-grams and dictionary need to match over the raw characters $\bm{a}$, while the type $n$-grams need to match over the sequence of character types $t(\bm{a}) := [t(a_0), t(a_1, ), ..., t(a_{N-1})]$.
This pattern matching is solved efficiently by using the \emph{Aho-Corasick (AC)} algorithm \cite{10.1145/360825.360855,10.3115/991886.991921}, which is also used with KyTea.
The AC algorithm uses the \textit{pattern matching automaton} (PMA), which performs in $O(N + \mathrm{occ})$ time in the most efficient cases, where $\mathrm{occ}$ is the number of pattern occurrences in the text.

The complexity of the AC algorithm additionally depends on the data structure of the inner PMA \cite{nieminen2007efficient}.
KyTea uses a binary search to discover state transitions of the PMA due to the large alphabet size of the Japanese characters, which requires $O(N \log \sigma + \mathrm{occ})$, where $\sigma$ is the expected number of possible transitions.
To mitigate this problem, we use \emph{compacted double-arrays} (CDAs) \cite{aoe1989efficient,10.1016/j.ipm.2006.04.004}. CDAs adjust the assignment of state IDs to share the memory space of their transition mappings as much as possible, which enables direct lookup of state transitions by character IDs so that the whole performance of the PMA goes back to $O(N + \mathrm{occ})$.

\subsection{Merging Character $n$-gram Scores}
\label{sec:algo::char}

\begin{figure}[t]
    \centering
    \includegraphics[width=0.95\hsize]{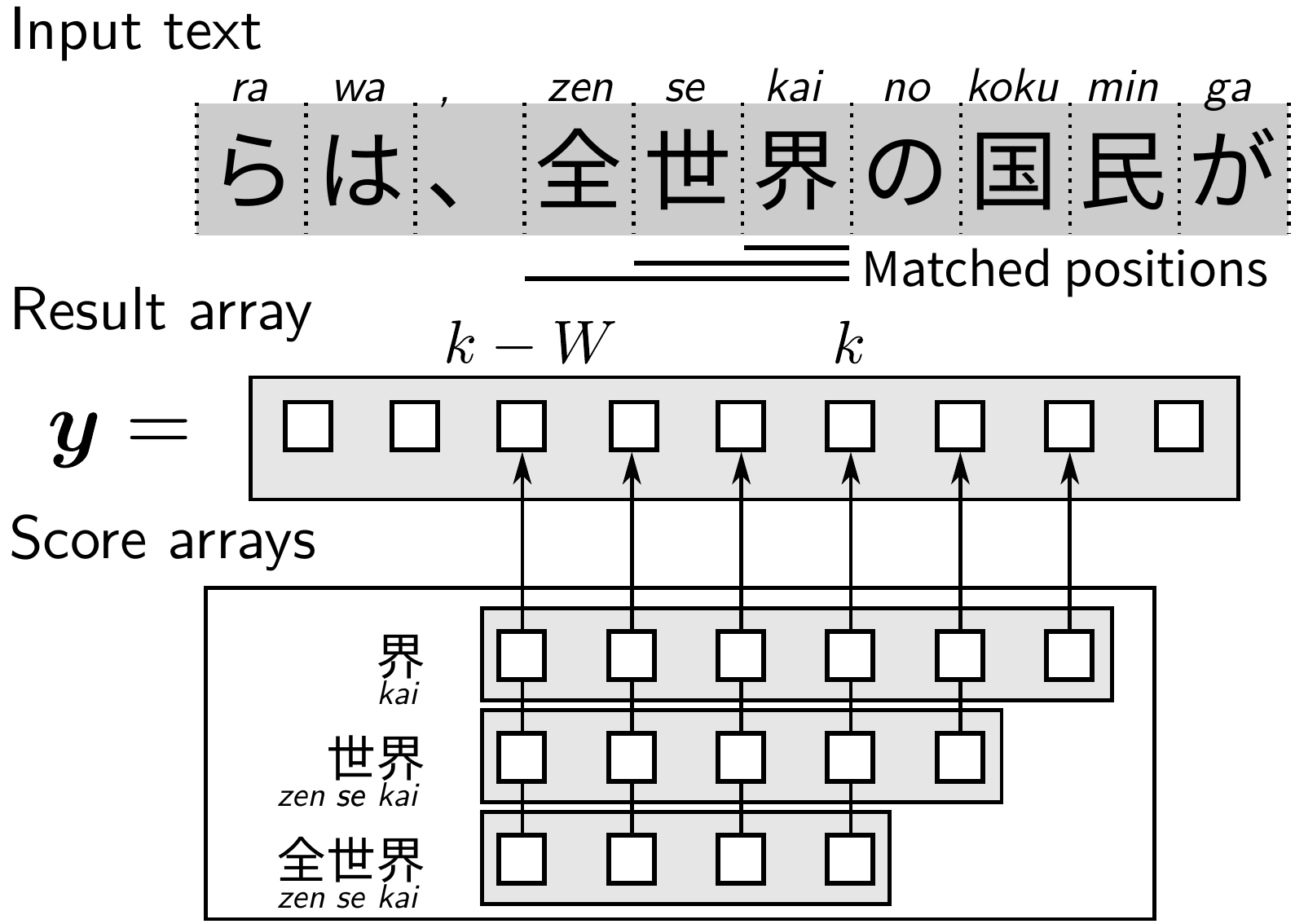}
    \caption{Integrating character $n$-gram scores to result array $\bm{y}$. $W=3$.
    $\bm{w}_{\textrm{pattern}}($``\Ja{\ruby[S]{界}{\textit{kai}}}''$)$ has 6 weights,
    $\bm{w}_{\textrm{pattern}}($``\Ja{\ruby[S]{世}{\textit{se}}\ruby[S]{界}{\textit{kai}}}''$)$ has 5 weights, and $\bm{w}_{\textrm{pattern}}($``\Ja{\ruby[S]{全}{\textit{zen}}\ruby[S]{世}{\textit{se}}\ruby[S]{界}{\textit{kai}}}''$)$ has 4 weights, as formulated in Equation (\ref{eq:w-pattern}).
    Each score array is integrated to position $k-W$ on $\bm{y}$, where $k$ is rightmost position of pattern.}
    \label{fig:weightvec}
\end{figure}

\begin{figure}[t]
    \centering
    \includegraphics[width=0.85\hsize]{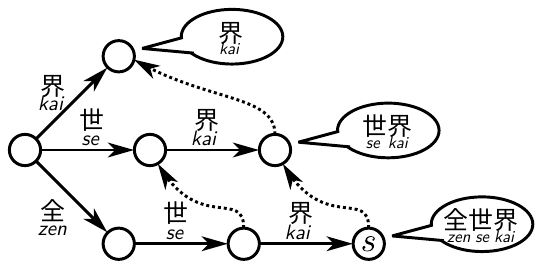}
    \caption{PMA built from three patterns (``\Ja{\ruby[S]{界}{\textit{kai}}}'', ``\Ja{\ruby[S]{世}{\textit{se}}\ruby[S]{界}{\textit{kai}}}'', ``\Ja{\ruby[S]{全}{\textit{zen}}\ruby[S]{世}{\textit{se}}\ruby[S]{界}{\textit{kai}}}''). Balloons indicate patterns reported at corresponding states. Dotted lines indicate failure edges to non-root states.}
    \label{fig:pma}
\end{figure}

PLC models usually introduce character $n$-grams with different $n$-s to capture a variety of contexts around the boundary. In this case, a longer $n$-gram overlaps several substring $n$-grams, and the PMA may report all possible $n$-grams sharing the same suffix as well as the longest $n$-gram.

Figure \ref{fig:weightvec} shows an example of summing three score arrays of ``\Ja{\ruby[S]{界}{\textit{kai}}}'', ``\Ja{\ruby[S]{世}{\textit{se}}\ruby[S]{界}{\textit{kai}}}'', and ``\Ja{\ruby[S]{全}{\textit{zen}}\ruby[S]{世}{\textit{se}}\ruby[S]{界}{\textit{kai}}}''. Since these patterns are suffixes of the longest one, summing the score array $\bm{w}_{\mathrm{pattern}}($``\Ja{\ruby[S]{全}{\textit{zen}}\ruby[S]{世}{\textit{se}}\ruby[S]{界}{\textit{kai}}}''$)$ always involves summing $\bm{w}_{\mathrm{pattern}}($``\Ja{\ruby[S]{世}{\textit{se}}\ruby[S]{界}{\textit{kai}}}''$)$ and $\bm{w}_{\mathrm{pattern}}($``\Ja{\ruby[S]{界}{\textit{kai}}}''$)$.

Figure \ref{fig:pma} shows a PMA built from ``\Ja{\ruby[S]{界}{\textit{kai}}}'', ``\Ja{\ruby[S]{世}{\textit{se}}\ruby[S]{界}{\textit{kai}}}'', and ``\Ja{\ruby[S]{全}{\textit{zen}}\ruby[S]{世}{\textit{se}}\ruby[S]{界}{\textit{kai}}}''. When the PMA reaches state $s$, it yields a list of all possible suffixes collected by tracing a chain of failure edges from $s$, and Algorithm \ref{alg:fast-pointwise-prediction} eventually aggregates all score arrays corresponding to the yielded suffixes. Since each state in the PMA always yields the same list of possible suffixes, we can calculate a \textit{partial sum} of the score arrays $\bm{w}_{\mathrm{state}}(s)$ in advance by summing over all possible $\bm{w}_{\mathrm{pattern}}(\cdot)$:
\begin{align}
\boldsymbol{w}_{\textrm{state}}(s) := \sum_{\boldsymbol{q} \in \mathcal{S}(s)} \boldsymbol{w}_{\textrm{pattern}}(\boldsymbol{q}),
\end{align}
where $\mathcal{S}(s)$ is the set of possible suffixes at $s$.
Using $\bm{w}_{\mathrm{state}}(\cdot)$ instead of $\bm{w}_{\mathrm{pattern}}(\cdot)$ improves the runtime efficiency by enabling Algorithm \ref{alg:fast-pointwise-prediction} to aggregate score arrays only once for each character boundary.

\subsection{Integrating Dictionary Words}
\label{sec:algo::dict}

\begin{figure}[t]
    \centering
    \includegraphics[width=0.95\hsize]{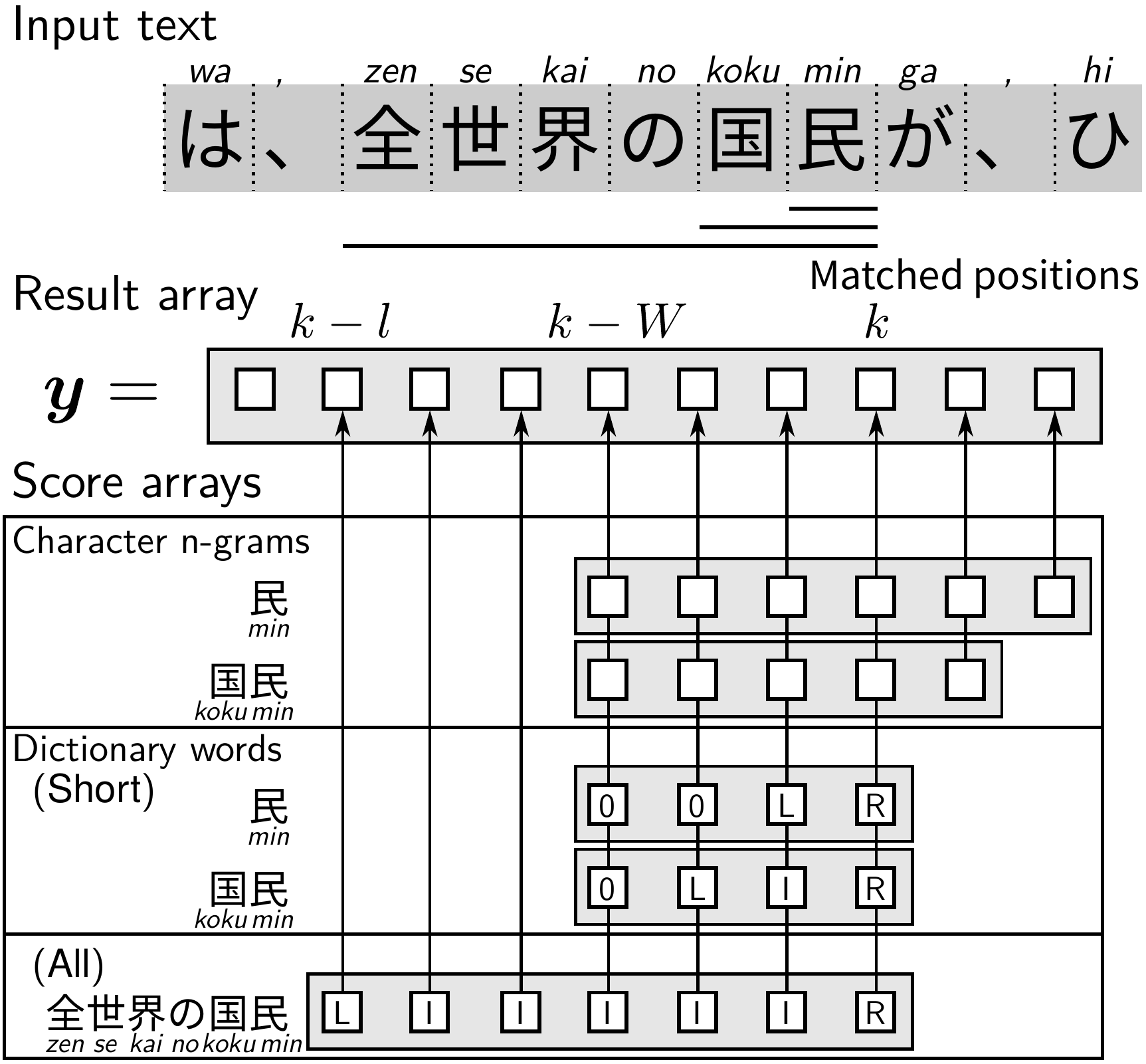}
    \caption{Difference between adding positions of character $n$-gram scores and dictionary word scores. $W=3$. L, I, and R indicate types of dictionary features. 0 indicates padding.}
    \label{fig:variable-offset}
\end{figure}

As discussed in Section \ref{sec:algo::matching}, PMAs of the character $n$-grams and dictionary run over the same sequence $\bm{a}$, and some of their patterns may overlap. This observation suggests that we can further integrate score arrays of dictionary words into the partial sum of the character $n$-gram scores to reduce a certain amount of burdens caused by matching dictionary words.
Therefore, our approach uses two methods to integrate dictionary features, as discussed in the following sections.

\subsubsection{Integrating Short Dictionary Words}
\label{sec:algo::dict::short}

As shown in Figure \ref{fig:weightvec}, the score arrays of character $n$-grams at position $k$ start with position $k-W$, and dictionary features of any words of lengths $l \leq W$ can be integrated into the partial sums of character $n$-grams with zero paddings (see Figure \ref{fig:variable-offset}). On the basis of this observation, it may be reasonable to integrate only short words while long words (of lengths $l > W$) remain in the separate PMA.
We call this method \textsf{Short}.

\subsubsection{Integrating All Dictionary Words}
\label{sec:algo::dict::all}

We can further consider integrating every dictionary word into the partial sums to eliminate the PMA of dictionary words.
Since score arrays of long words cover beyond the range of character $n$-gram scores (as in the last case of Figure \ref{fig:variable-offset}), we need to prepare additional arrays to store partial sums for all long words. We also need to determine the correct starting positions of the score summation, which may incur an additional cost for the calculation.
We call this method \textsf{All}.

\subsection{Caching Type $n$-gram Scores}
\label{sec:algo::type}

Type $n$-gram scores are calculated similarly to character $n$-grams, but their alphabet size is limited. As described in Section \ref{sec:pointwise}, KyTea models distinguish only six character types. Since the sliding window of size $W$ contains a sequence of character types of length $2W$, the number of possible type $n$-gram sequences is only $6^{2W}$. This is small enough to store all resulting scores of possible sequences at initialization for a reasonable $W$, typically $W = 3$.\footnote{In accordance with the analysis in Appendix \ref{sec:ex::nw-size}.} This approach allows the score calculation of type $n$-grams by looking up only one integrated score, avoiding the PMA of type $n$-grams and summing corresponding score arrays.

\begin{figure}[t]
    \centering
    \includegraphics[width=0.95\hsize]{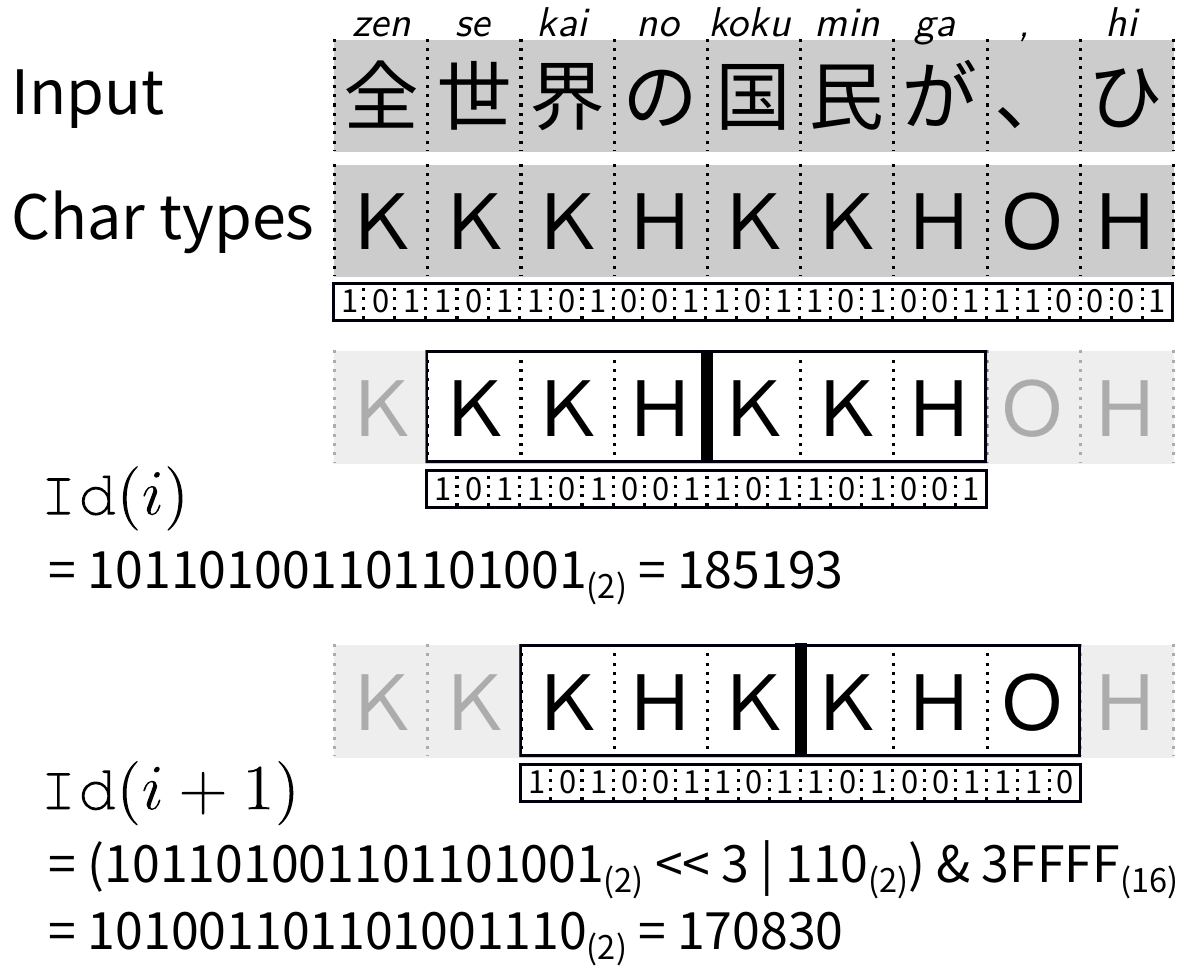}
    \caption{Calculating sequence IDs of character types with $W = 3$. Binary sequences under character types indicate codes related to each character type: for example, $\CodeFn{\textrm{H}}=001_{(2)}$. The sequence ID of next sliding window can be calculated by current sequence ID and incoming character type.}
    \label{fig:type-weights}
\end{figure}

Specifically, we assign a binary code of 3 bits $\CodeFn{t} \in [1, 6]$ for each character type $t$ and define the \textit{sequence ID} $\IdFn{i}$ as follows:
\begin{align}
& \IdFn{i} := \sum_{k=0}^{2W-1} 2^{3(2W-1-k)} c(i-W+k), \\
& c(i) := \left \{
    \begin{array}{ll}
    \CodeFn{t_i} & \textrm{if} \ i \in [0, N-1], \\
    0 & \textrm{otherwise}, \\
    \end{array}
\right .
\end{align}
for each sliding window at position $i$, where $t_i := t(a_i)$. $\IdFn{i}$ is a $6W$ bit integer used as the address of the integrated type $n$-gram score.
Since $\IdFn{i}$ shares most of its subsequence with the neighboring window $\IdFn{i+1}$, as shown in Figure \ref{fig:type-weights}, we can induce $\IdFn{\cdot}$ values recurrently as follows:
\begin{align}
& \IdFn{-W} := 0, \\
& \IdFn{i+1} := (2^3 \IdFn{i} + c(i+W)) \% 2^{6W},
\end{align}
where $\%$ indicates the modulo operation. This calculation can be executed by only a few bit operations and requires a constant time complexity for each character boundary.
\section{Experiments}

\begin{table*}[t]
\centering
\small
\caption{Average time elapsed to tokenize test data [ms]}
\label{tab:speed}
\begin{tabular}{l|rrr}
\hline
Tokenizer & Time & Time $-$ (a) & SD \\ \hline \hline
KyTea (2020-04-03) & 136.6 & & 5.6 \\ \hline
\textbf{Our methods} & & &  \\
\hspace{1em} (a): \S\ref{sec:algo::matching} Efficient Pattern Matching & 33.2 & (0) & 0.5 \\
\hspace{1em} (b): (a) $+$ \S\ref{sec:algo::char} Merging Character $n$-gram Scores & 31.2 & -2.0 & 0.6 \\
\hspace{1em} (c): (a) $+$ \S\ref{sec:algo::dict} Integrating Dictionary Scores (\textsf{All}) & 29.4 & -3.8 & 0.5 \\
\hspace{1em} (d): (a) $+$ \S\ref{sec:algo::type} Caching Type $n$-gram Scores & 30.2 & -3.0 & 0.5 \\
\hspace{1em} (e): (a) $+$ \S\ref{sec:algo::char} $+$ \S\ref{sec:algo::dict} (\textsf{All}) $+$ \S\ref{sec:algo::type} & \textbf{23.8} & \textbf{-9.4} & 0.4 \\ \hline \hline
MeCab (2020-09-14) & &  &  \\
\hspace{1em} IPADic & 65.4 & & 1.9 \\
\hspace{1em} UniDic & 161.5 & & 7.3 \\ \hline
sudachi.rs (0.6.4-a1) & 169.5 & & 3.4 \\ \hline
\end{tabular}
\end{table*}

\begin{table*}[t]
\centering
\small
\caption{Average number of score arrays aggregated during tokenizing test data {[$\times 10^3$]}}
\label{tab:breakdown:match}
\begin{tabular}{l|rrrrrr} \hline
Subroutine & \multicolumn{1}{c}{(a)} & \multicolumn{1}{c}{(b)} & \multicolumn{1}{c}{(a)$+$\textsf{Short}} & \multicolumn{1}{c}{(a)$+$\textsf{All}} & \multicolumn{1}{c}{(b)$+$\textsf{Short}} & \multicolumn{1}{c}{(b)$+$\textsf{All}} \\ \hline\hline
Character $n$-grams & 360. & 203. & 386. & 392 & 204. & 204. \\
Dictionary words & 307. & 307. & 6. & -- & 6. & -- \\ \hline\hline
Total & 667. & 509. & 392. & 392 & 210. & 204. \\ \hline
\end{tabular}
\end{table*}

\begin{table*}[t]
\centering
\small
\caption{Average time elapsed to process character $n$-grams and dictionary [ms]}
\label{tab:breakdown:time}
\begin{tabular}{l|rrrrrr} \hline
Subroutine & \multicolumn{1}{c}{(a)} & \multicolumn{1}{c}{(b)} & \multicolumn{1}{c}{(a)$+$\textsf{Short}} & \multicolumn{1}{c}{(a)$+$\textsf{All}} & \multicolumn{1}{c}{(b)$+$\textsf{Short}} & \multicolumn{1}{c}{(b)$+$\textsf{All}} \\ \hline\hline
Character $n$-grams &  &  &  & &  &  \\
\hspace{1em} Pattern matching & 6.15 & 5.99 & 7.68 & 9.65 & 7.35 & 9.09 \\
\hspace{1em} All & 7.46 & 6.75 & 10.96 & 17.75 & 9.60 & 15.88 \\ \hline
Dictionary words &  &  &  & &  &  \\
\hspace{1em} Pattern matching & 8.28 & 8.26 & 5.82 & -- & 5.74 & -- \\
\hspace{1em} All & 13.63 & 13.56 & 6.27 & -- & 6.36 & -- \\ \hline\hline
All subroutines & 23.67 & 22.82 & 20.77 & 17.75 & 19.40 & 15.88 \\ \hline
\end{tabular}
\end{table*}

\subsection{Setup}

We evaluated the tokenization speed of our methods with multiple combinations and compared them with conventional tokenization methods, i.e. KyTea, MeCab, and sudachi.rs, which are explained in the next subsection.
We used short unit words (SUWs) in the BCCWJ 1.1 corpus \cite{10.1007/s10579-013-9261-0}\footnote{\url{https://clrd.ninjal.ac.jp/bccwj/en/}} to train PLC models. The corpus consists of 60k Japanese sentences with manually annotated SUW boundaries. 
We also used 667k words for dictionary features extracted from UniDic 3.1.0 \cite{den2007unidic}\footnote{\url{https://clrd.ninjal.ac.jp/unidic/en/} (GPL-2 or LGPL-2.1 or BSD-3)} with manual filtering.\footnote{We removed all words containing whitespaces.}
Our methods were implemented in Rust and compiled by rustc 1.60.0 with optimization flag \texttt{opt-level=3}. The other methods were compiled by GCC 11.2.0 or the same rustc with their recommended configuration.
For each experiment, we conducted 10-fold cross-validation: nine fractions for training and the remaining one for test. The PLC model was trained by LIBLINEAR with L1 regularization \cite{tibshirani-1996-lasso}. The same PLC model was used in both KyTea and our methods.
We fixed several hyperparameters to obtain representative metrics of each method: the penalty parameter $C$ to 1, $W$ to 3, and maximum length of $n$-grams to 3, in accordance with the analysis in Appendix \ref{sec:accuracy} and \newcite{mori2011pointwise}.

We mainly compared the performance of the methods on a long-lived server service.
Therefore, we measured only the computing overhead imposed by the actual tokenization processes.
Specifically, model initialization (loading the parameters and preprocessing integrated scores) was omitted from the resulting measure because they are used only once during the runtime. In addition, all test sentences were loaded onto the memory in advance to avoid constant disk access.

We conducted all experiments in a single thread on a dedicated machine with an Intel Core i7-8086K CPU (4GHz, 6 cores, 32KiB L1, 256KiB L2, 12MiB L3) and 64GiB DDR4 RAM.

\subsection{Baseline Methods}
\label{sec:ex:baselines}

We introduced the following methods as our baselines:

\paragraph{KyTea\footnote{\url{https://github.com/neubig/kytea} (Apache-2.0)}}
The original implementation of the PLC method.
We verified that all confidence scores of character boundaries calculated with KyTea and our implementation are identical\footnote{Models are quantized by the design of both KyTea and our methods, so all calculations were done by integers.} except for dictionary features.\footnote{Dictionary features are slightly changed from KyTea to design Method \S\ref{sec:algo::dict}, but this change has only a trivial impact on tokenization accuracy, as shown in Appendix \ref{sec:dict-diff}.}

\paragraph{MeCab\footnote{\url{https://github.com/taku910/mecab} (GPL-2 or LGPL-2.1 or BSD-3)}}
A widely used Japanese lattice-based tokenization method.
We used the IPADic dictionary \cite{ipadic} as well as UniDic.\footnote{We provide a reference performance of MeCab with a typical configuration (IPADic) for a fair comparison of speeds.}

\paragraph{sudachi.rs\footnote{\url{https://github.com/WorksApplications/sudachi.rs} (Apache-2.0)}}
An efficient implementation of Sudachi \cite{takaoka-etal-2018-sudachi} and is a lattice-based method. We selected Sudachi because it is widely used as an internal tokenizer of larger systems such as spaCy.\footnote{\url{https://spacy.io/}}
We used the \textit{SudachiDict-core 20210802} model and disabled all post-processing.

\subsection{Overall Speed Comparison}
\label{sec:ex::comp}

Table \ref{tab:speed} shows the average time elapsed to tokenize the test data. The results from KyTea and our methods include only time elapsed by the boundary classification, while MeCab and sudachi.rs involve the complete morphological analysis that is difficult to separate due to the model formulation.\footnote{Lattice methods are designed to take into account a joint distribution of tokenization and morphology.}

First, we focus on five different settings (a) to (e) of our methods described in Section \ref{sec:algo}.
The settings (b), (c), and (d) run faster than (a), demonstrating that these preprocessing approaches are practical for suppressing computation time. We can also see that applying all preprocessing (e) achieves the fastest result, and the overall improvement is comparable with the sum of (b), (c), and (d). This result suggests that these techniques are orthogonal and improve different parts of computation in the whole algorithm.

Comparing (e) and other tools, our method achieves 5.7 times faster than KyTea, 6.8 times faster than MeCab with the same dictionary,\footnote{In Japanese, there are multiple definitions of ``words''. PLC models rely on word boundary labels annotated in the training corpus (BCCWJ), and the word unit of the dictionary must be compatible with the corpus standard to avoid unnecessary lacking tokenization accuracy. Since the word unit of IPADic is not compatible with BCCWJ, we did not prepare the results of PLC models with IPADic.} and 7.1 times faster than sudachi.rs.

\subsection{Performance of Subroutines}
\label{sec:ex::detail}

Table \ref{tab:breakdown:match} shows the average number of score arrays aggregated during the score calculation of all test examples. We focused on the number of arrays rather than the number of actual scores because most score arrays can be aggregated by at most a few SIMD instructions. We do not show corresponding metrics of type $n$-grams because its calculation was eliminated by the method described in Section \ref{sec:algo::type}.

We can see that merging character $n$-gram scores (b) reduces 44\% of array summations involved by character $n$-gram features, and integrating short dictionary words (a)$+$\textsf{Short} reduces 98\% of array summations involved by dictionary features compared to (a). (a)$+$\textsf{Short} also slightly increases the number of array summations in character $n$-grams due to introducing unseen character $n$-grams derived from the dictionary.

In contrast, combining both methods, (b)$+$\textsf{Short} successfully reduced the calculation of character $n$-gram scores to a comparable range of (b). This tendency shows that even if the dictionary contains unseen character $n$-grams, they can be integrated into the partial sums of other patterns in most cases.
Comparing methods \textsf{Short} and \textsf{All} in both (a) and (b), we can see no significant difference in the number of score summations by integrating long words.

Table \ref{tab:breakdown:time} shows the average time elapsed during each subroutine.
We measured each metric by simply disabling other subroutines from the whole process. The last row shows the time with all subroutines, which should be longer than the sum of all individual metrics due to the lack of caching efficiency.
We can see that the time of character $n$-grams in (b)$+$\textsf{All} is longer than that in (b)$+$\textsf{Short} nevertheless, the numbers of score summations in Table \ref{tab:breakdown:match} are almost the same.
This is expected because the method \textsf{All} introduces additional complexity, as discussed in Section \ref{sec:algo::dict::all}.
However, we can also see that the whole process of (b)$+$\textsf{All} achieved faster performance than (b)$+$\textsf{Short} because (b)$+$\textsf{All} eventually removes the whole process of dictionary features completely.

\subsection{Effect of Model Size}
\label{sec:ex::l1}

\begin{figure}[t]
    \centering
    \includegraphics[width=\hsize]{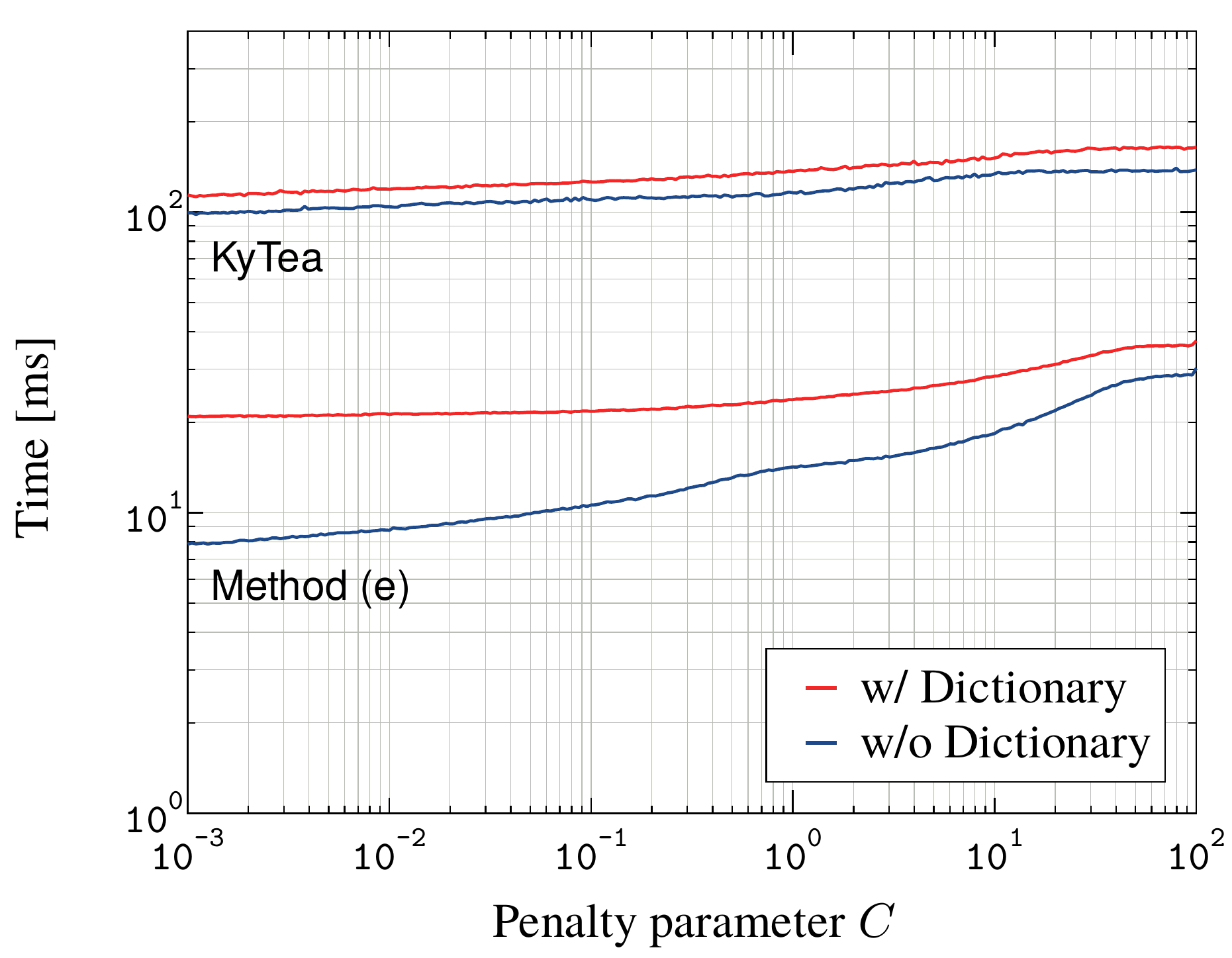}
    \caption{Effect of penalty parameter $C$ against elapsed time of each method}
    \label{fig:elapsed-time-vaporetto}
\end{figure}

\begin{figure}[t]
    \centering
    \includegraphics[width=\hsize]{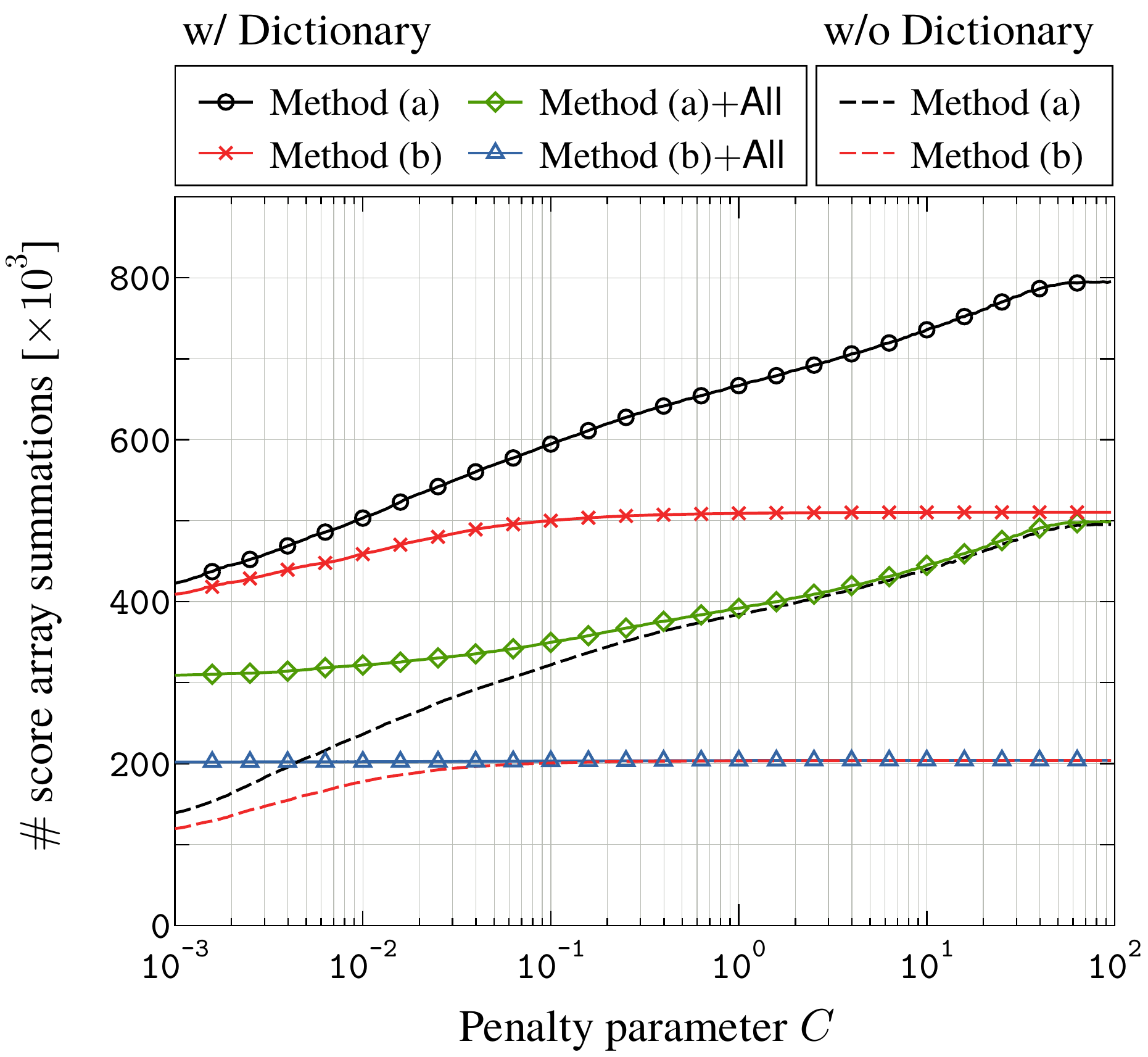}
    \caption{Effect of $C$ against number of scorearray summations}
    \label{fig:pattern-occ}
\end{figure}

\begin{figure}[t]
    \centering
    \includegraphics[width=\hsize]{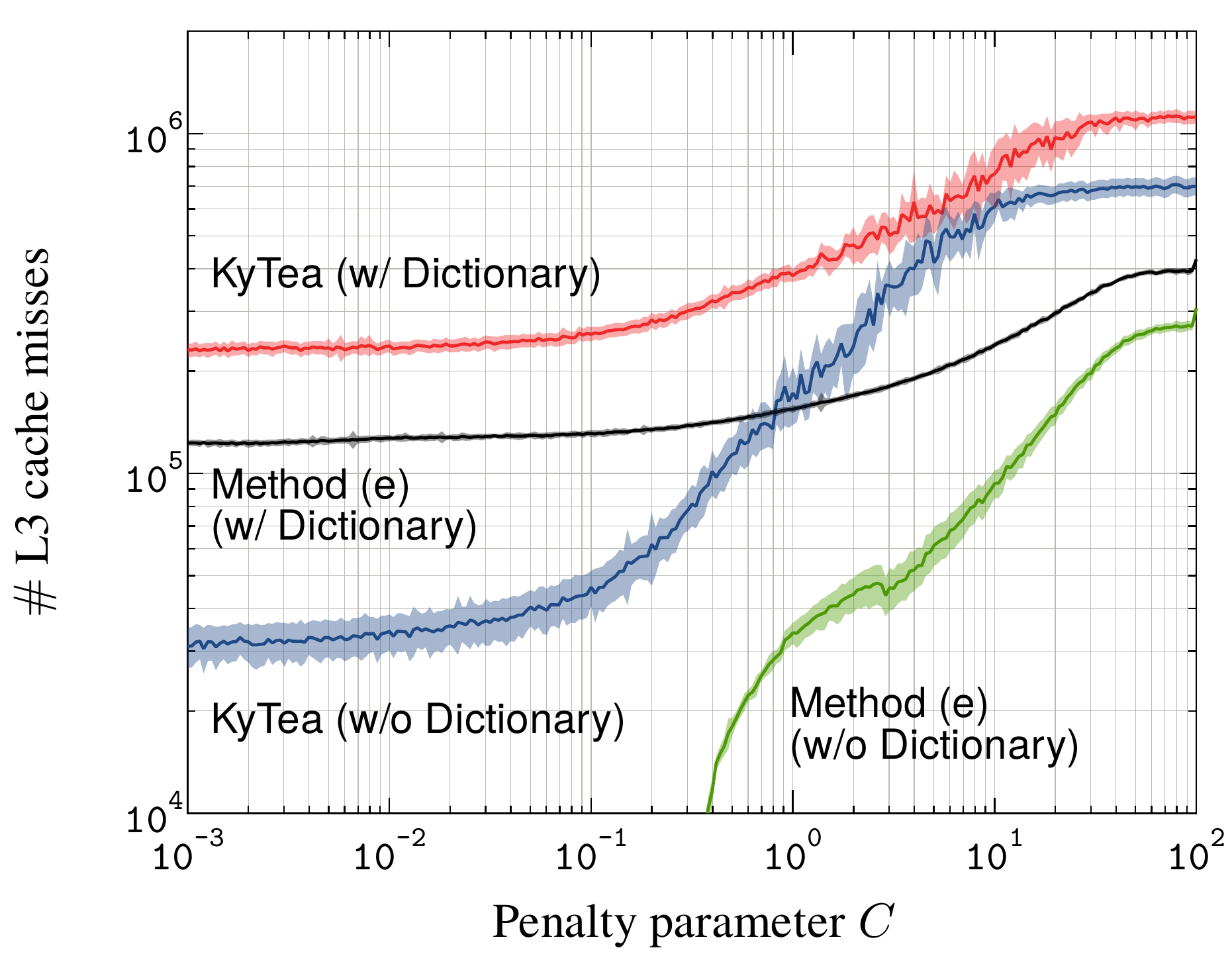}
    \caption{Effect of $C$ against number of L3 cache misses of each method. Lines and areas indicate means and standard deviations, respectively.}
    \label{fig:cache}
\end{figure}

\begin{figure}[t]
    \centering
    \includegraphics[width=\hsize]{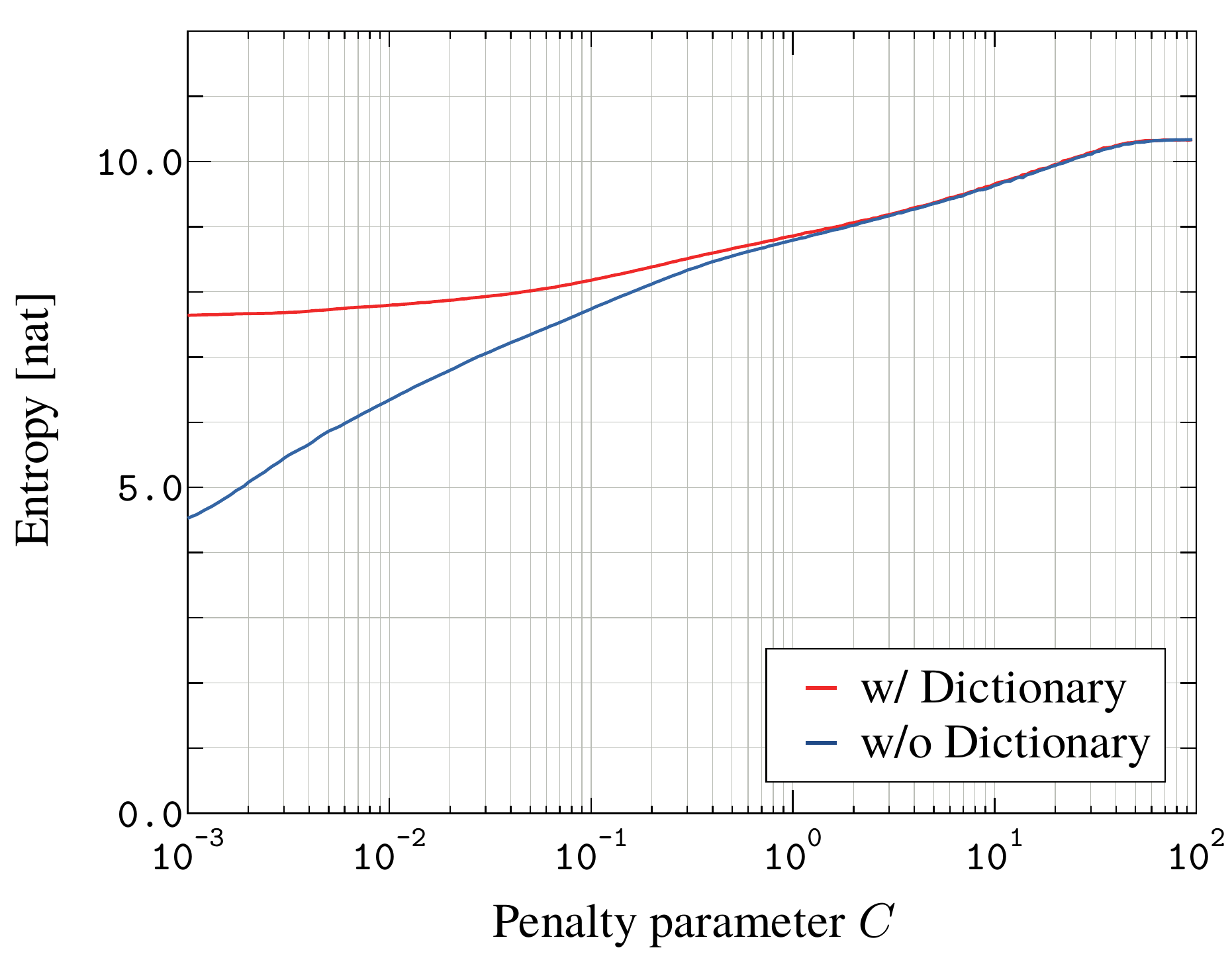}
    \caption{Entropy of access-frequency distribution of character $n$-gram score arrays with (e)}
    \label{fig:score-array-entropy}
\end{figure}

We analyzed the impact of model sizes against tokenization efficiency by varying the penalty parameter $C$ of L1 regularization \cite{tibshirani-1996-lasso} when training SVMs.\footnote{We followed the same definition of $C$ in \newcite{10.5555/1390681.1442794}: the smaller it is, the stronger the regularization. The actual relation between model sizes and $C$ is shown in Appendix \ref{sec:model-size}.}
L1 regularization squashes a certain number of weights into 0, and such weights can be removed from the model \cite{gao-etal-2007-comparative}.

Figure \ref{fig:elapsed-time-vaporetto} shows the whole processing time of the method (e) and KyTea with varying $C$.
Intuitively, both models achieve faster processing speed with small $C$ (strong regularization), and our method achieves better processing speed against KyTea for every $C$.

Considering the gradient of each plot in Figure \ref{fig:elapsed-time-vaporetto}, we can also see that $C$ brings a larger exponential effect against the processing speed of the method (e) than KyTea. This tendency is reasonable because our methods heavily rely on the CPU architecture: large models may cause many disruptions to efficient computing. We provide the details of several observations of this tendency as follows.

We hypothesized that this tendency is caused by increasing the number of array summations because a larger model may discover more patterns on the same input.
As Figure \ref{fig:pattern-occ} shows, this hypothesis is not correct because merging character $n$-grams ((b) and (b)$+$\textsf{All}) effectively suppressed the number of array summations under a certain upper bound even when $C$ was large (weak regularization). This means that the algorithm's complexity does not reduce speed on large $C$.

We also investigated the hardware-level efficiency of each method. As clearly shown in Figure \ref{fig:cache}, the number of CPU cache misses\footnote{Counted by the \texttt{perf\_event\_open} system call on Linux.} increased significantly by increasing $C$ despite maintaining the number of summation operations.
This phenomenon is explained by observing the access-frequency of score arrays. Figure \ref{fig:score-array-entropy} summarizes the access frequency distribution of score arrays as entropy and shows that large models require more varied score arrays than small ones to calculate the final scores. Since the CPU cache can remember only the neighboring contents around the memory accessed recently, requiring various parts of memory lacks utilization of the cache, resulting in an overall speed reduction of the algorithm.
\section{Conclusion and Future Work}

We introduced multiple techniques to improve the efficiency of the Japanese tokenizer based on Pointwise Linear Classification (PLC) models. Experiments clearly showed substantial improvements brought by our methods compared with the baseline implementation (KyTea) and other tokenization tools in terms of tokenization speed.
Although we focused especially on the tokenization task, some of the techniques presented in this paper are generic and can also be applied to other tasks if we can decompose them into a sequence of multiple problems in the same manner.

We improved only the tokenization speed in this study because it is an essential part of practical use-cases of tokenizers. Improving the overall efficiency of PLC-based lexical analysis is also challenging; it is one of the main focuses of our future work. For example, PLC-based part-of-speech tagging requires a much larger alphabet size (set of \textit{words} rather than \textit{characters}), and further improvement of the PMA architecture is required.
\section{Limitations}

Since our data structures are specialized to Japanese, caching type $n$-gram scores described in Section 3.4 can not be straightforwardly used in other languages. More precisely, the method exploits the characteristic that the number of Japanese character types is 6. In other languages, such as Chinese and Korean, modifications are required to use this idea.

\bibliography{anthology,custom}
\bibliographystyle{acl_natbib}

\clearpage

\appendix

\begin{table*}[t]
\centering
\caption{Effect of window size $W$ and $n$-gram size $n$ on elapsed time [ms] and accuracy}
\label{tab:windowsize}
\begin{tabular}{c|rcc|rcc}
\hline
\multirow{2}{*}{$W,n$} & \multicolumn{3}{c}{SUW} & \multicolumn{3}{|c}{LUW} \\
& \multicolumn{1}{c}{Time} & \multicolumn{1}{c}{$F_1$} & \multicolumn{1}{c}{Error rate} & \multicolumn{1}{|c}{Time} & \multicolumn{1}{c}{$F_1$} & \multicolumn{1}{c}{Error rate} \\ \hline \hline
1,1 & 7.7 & 0.8265 & 0.0887 & 7.6 & 0.8207 & 0.0749 \\ \hline
2,2 & 11.3 & 0.9810 & 0.0093 & 11.1 & 0.9608 & 0.0155 \\ \hline
3,3 & 13.3 & 0.9867 & 0.0066 & 13.3 & 0.9779 & 0.0088 \\ \hline
4,4 & 15.7 & 0.9867 & 0.0066 & 15.9 & 0.9805 & 0.0078 \\ \hline
5,5 & 20.1 & 0.9862 & 0.0069 & 20.5 & 0.9807 & 0.0077 \\ \hline
6,6 & 23.3 & 0.9857 & 0.0071 & 24.2 & 0.9804 & 0.0078 \\ \hline
7,7 & 25.6 & 0.9850 & 0.0074 & 27.1 & 0.9799 & 0.0080 \\ \hline
\end{tabular}
\end{table*}

\section{Appendix}
\label{sec:appendix}

\subsection{Differences in Dictionary Features}
\label{sec:dict-diff}

We slightly changed dictionary features from KyTea to design the methods described in Section \ref{sec:algo::dict}.
Specifically, KyTea uses binary features indicating the relation of dictionary words and character boundaries. In contrast, our implementation uses frequency features depending on how many dictionary words overlap.
Table \ref{tab:dict-diff} shows the effect of changing dictionary features against tokenization accuracy.
We can see that the change in dictionary features has only a trivial impact on tokenization accuracy.

\begin{table}[t]
\centering
\caption{Effect of changing dictionary features on accuracy}
\label{tab:dict-diff}
\begin{tabular}{l|c|c}
\hline
Tokenizer & $F_1$ & Error rate \\ \hline \hline
KyTea & 0.9933 & 0.0033 \\ \hline
Our method & 0.9934 & 0.0032 \\ \hline
\end{tabular}
\end{table}

\subsection{Accuracy Metrics}

To measure tokenization accuracy, we choose two metrics:
\textit{boundary error rate}, which is the ratio of false classifications for all character boundaries,
and the \textit{word-wise $F_1$ measure} \cite{nagata-1994-stochastic}.

\subsection{Relation between Penalty Parameter and Accuracy}
\label{sec:accuracy}

\begin{figure}[t]
    \centering
    \includegraphics[width=\hsize]{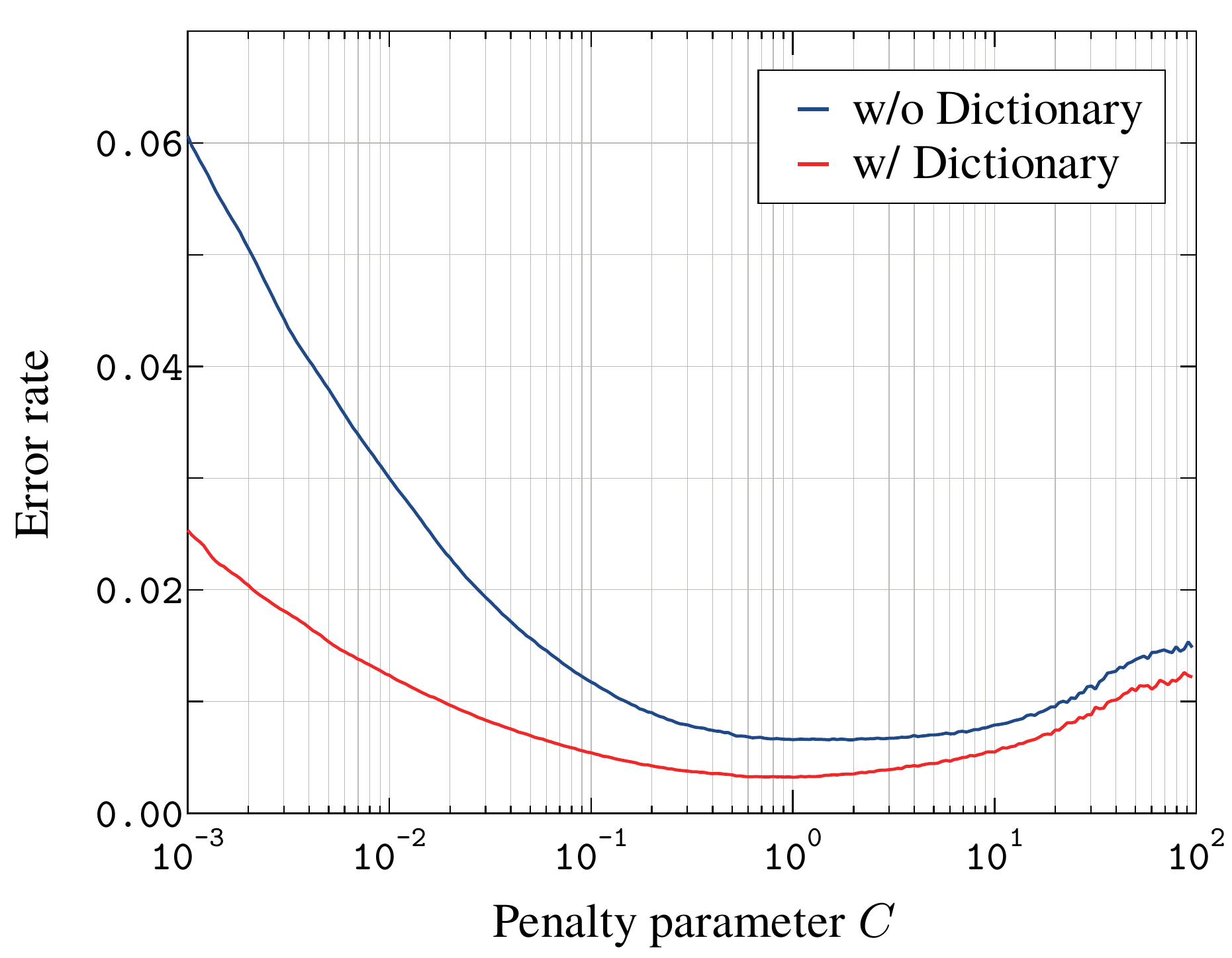}
    \caption{Effect of $C$ against mean error rate}
    \label{fig:error-rate}
\end{figure}

We investigated the effect of the strength of L1 regularization against tokenization accuracy.
Figure \ref{fig:error-rate} shows the tendency of the boundary error rate with varying $C$. As discussed in Section \ref{sec:ex:baselines}, our methods and KyTea share the same accuracy.
We can see that the error rate becomes minimum around $C=1$ with and without using a dictionary.

\subsection{Relation between Penalty Parameter and Model Size}
\label{sec:model-size}

\begin{figure}[t]
    \centering
    \includegraphics[width=\hsize]{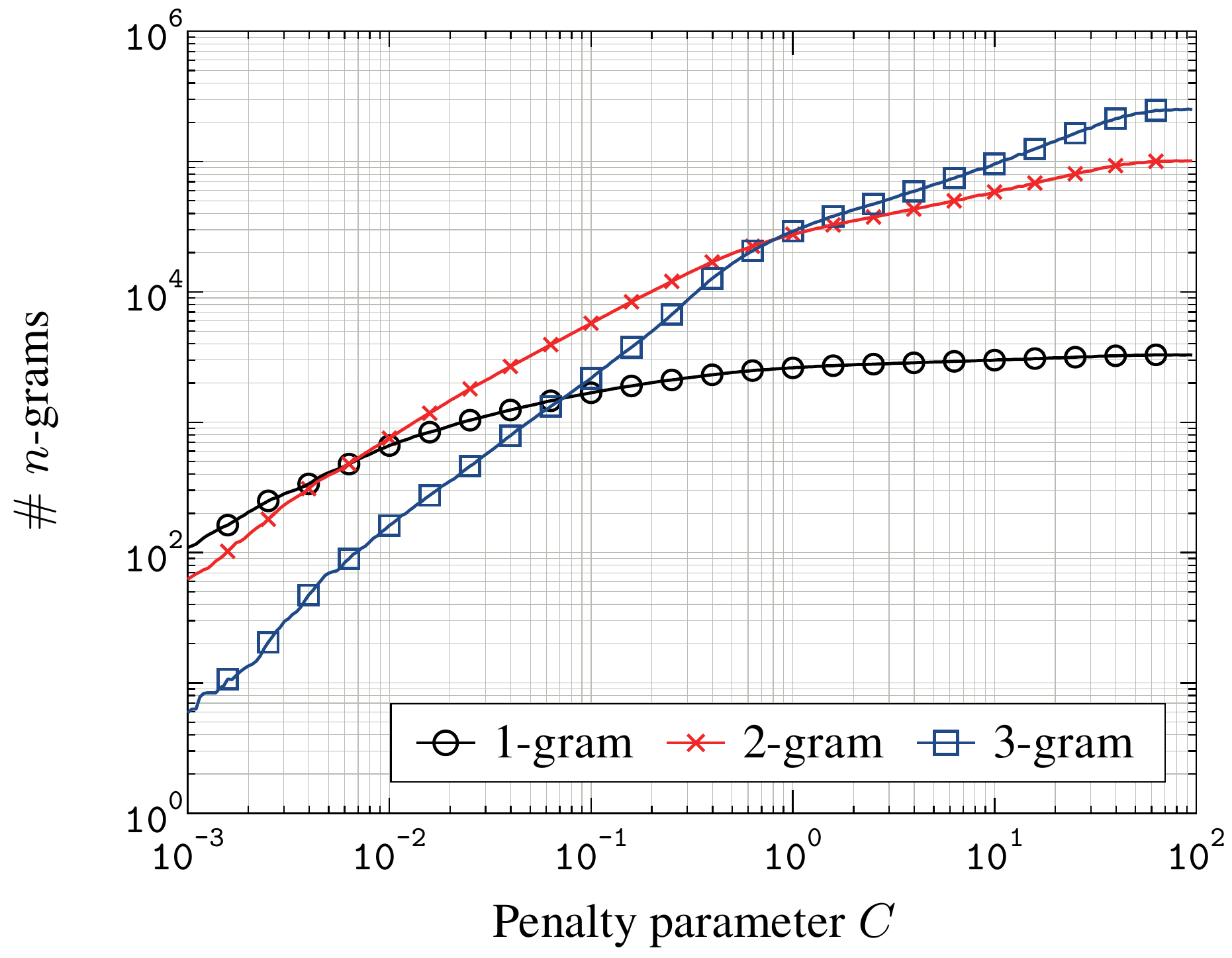}
    \caption{Effect of $C$ on number of $n$-grams in model (w/o Dictionary)}
    \label{fig:model-n-grams}
\end{figure}

\begin{figure}[t]
    \centering
    \includegraphics[width=\hsize]{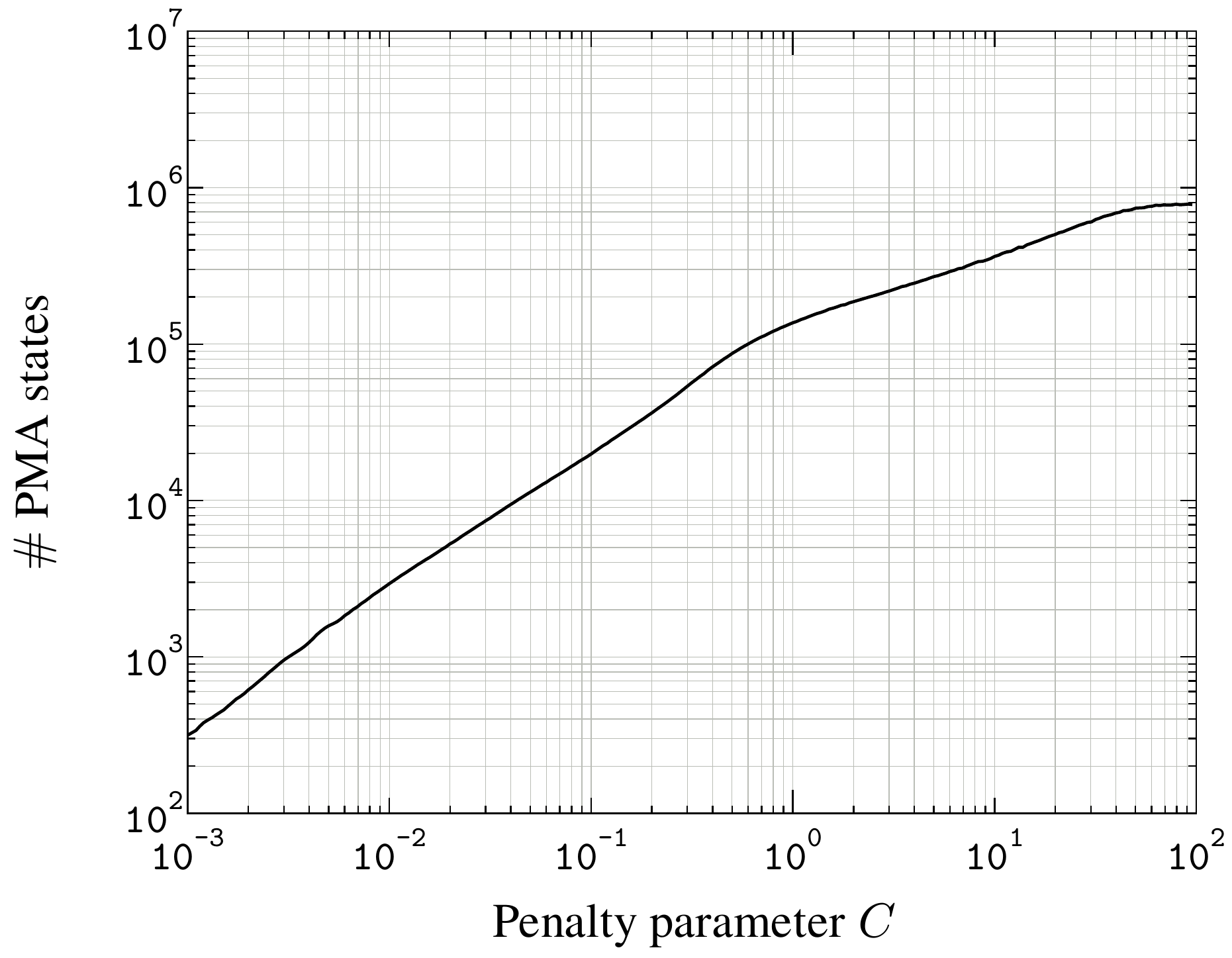}
    \caption{Effect of $C$ on number of PMA states in model (w/o Dictionary)}
    \label{fig:model-n-states}
\end{figure}

We investigated the effect of the strength of L1 regularization on model size.
Figure \ref{fig:model-n-grams} shows the tendency of $n$-grams in the model, and Figure \ref{fig:model-n-states} shows the number of states in the PMA with varying $C$.
We can see that larger $C$ (weak regularization) yields more $n$-grams with non-zero weights, requiring more PMA states. We can also see that $C$ has different effects for $n$-grams with different $n$-s.

\subsection{Relation between Window Size and Accuracy}
\label{sec:ex::nw-size}

We investigated the effect of $W$ and $n$ against tokenization speed and accuracy. We did not use any dictionary for this experiment because the dictionary feature is independent of both $W$ and $n$.
For models with $W \geq 4$, we did not introduce caching type $n$-gram scores discussed in Section \ref{sec:algo::type} because this requires a large amount of memory.

Table \ref{tab:windowsize} shows the comparison of tokenization accuracy and speed.
We can see that tokenization accuracy is saturated with certain $W$ and $n$ (3 for SUW and 5 for LUW), although the tokenization speed drops when we select larger $W$ and $n$.

\end{document}